\documentclass[conference]{IEEEtran}
\IEEEoverridecommandlockouts
\usepackage{cite}
\usepackage{amsmath,amssymb,amsfonts}
\usepackage{algorithmic}
\usepackage{graphicx}
\usepackage{textcomp}
\usepackage{xcolor}
\usepackage{fancyhdr}
\usepackage{float}
\usepackage{caption}
\usepackage{wrapfig}
\usepackage{lipsum}% generate text for the example
\usepackage{booktabs}
\usepackage{hyperref}

\def\BibTeX{{\rm B\kern-.05em{\sc i\kern-.025em b}\kern-.08em
    T\kern-.1667em\lower.7ex\hbox{E}\kern-.125emX}}
    
\fancypagestyle{firstpagefooter}{%
  \fancyhf{}
  
}

\begin{document}

\title{Comparison of Machine Learning Approaches for Classifying Spinodal Events}

\author{\IEEEauthorblockN{Ashwini Malviya}
\IEEEauthorblockA{\textit{Department of physics } \\
\textit{Indian Institute of Technology Roorkee}\\
Roorkee, India \\
a\_malviya@ph.iitr.ac.in}
\and
\IEEEauthorblockN{ Sparsh Mittal}
\IEEEauthorblockA{\textit{Department of Electronics and Communication Engineering} \\
\textit{Indian Institute of Technology Roorkee}\\
Roorkee, India  \\
sparsh.mittal@ece.iitr.ac.in}
\and
\IEEEauthorblockN{}
\IEEEauthorblockA{} \\
\textit{}
}

\maketitle

\begin{abstract}
In this work, we compare the performance of deep learning models for classifying the spinodal dataset. We evaluate state-of-the-art models (MobileViT, NAT, EfficientNet, CNN), alongside several ensemble models (majority voting, AdaBoost). Additionally, we explore the dataset in a transformed color space. Our findings show that NAT and MobileViT outperform other models, achieving the highest metrics—accuracy, AUC, and F1 score—on both training and testing data (NAT: 94.65, 0.98, 0.94; MobileViT: 94.20, 0.98, 0.94), surpassing the earlier CNN model (88.44, 0.95, 0.88). We also discuss failure cases for the top-performing models.
\end{abstract}

\begin{IEEEkeywords}
CNN, NAT, MobileVit, QCD, Spinodal 
\end{IEEEkeywords}

\thispagestyle{firstpagefooter}

\section{Introduction}
In the context of first-order phase transitions in Quantum Chromodynamics (QCD), Spinodal Decomposition refers to a dynamically unstable mechanism of phase transformation. In the phase diagram, it is represented by the spinodal region, where the key feature is the non-equilibrium process driven by the exponential growth of small perturbations, resulting in an unstable segment of the Equation of State (EOS). This phenomenon is characterized by the formation of domains or regions of fluctuations. In contrast, Maxwell's thermodynamic construction describes a first-order transition without instabilities, where the mixed-phase region exhibits constant pressure and phase coexistence throughout the transition.

Heavy-ion collisions, such as those occurring at the Relativistic Heavy Ion Collider or the Large Hadron Collider, create extreme conditions leading to the formation of Quark-Gluon Plasma (QGP) at high temperatures and densities. As the plasma cools through a first-order phase transition, mechanical instabilities and domain formations (the spinodal region) may arise \cite{Scavenius_2001,Randrup_2004}. Detecting these signatures can provide critical evidence about the nature of the phase transition \cite{Li_2017,PhysRevLett.109.212301,CHOMAZ2004263}.

The QCD phase diagram delineates various phases of nuclear matter with respect to temperature and chemical potential. It is anticipated that a first-order phase transition line in this diagram terminates at a critical point \cite{BZDAK20201}. In the vicinity of this critical point, the system can enter the spinodal region \cite{STEINHEIMER2021121867,Sasaki_2007}, underscoring the importance of understanding spinodal decomposition as a key phenomenon in phase separation dynamics. 

Hydrodynamic simulations, which model the space-time evolution of QGP resulting from heavy-ion collisions, benefit from incorporating spinodal decomposition into the models. This allows for a more accurate representation of phase transition dynamics, particularly in the unstable regions of the EOS \cite{Steinheimer_2013}.

Although spinodal decomposition is primarily studied in the context of heavy-ion collisions, it may also be relevant for dense matter in neutron star mergers, where extreme conditions could induce phase transitions in nuclear matter. Therefore, detecting spinodal decomposition is a critical task. Given the broad range of applications, this work explores machine learning models for the classification of phase transitions into two categories: spinodal and non-spinodal.

\section{Dataset}
The dataset used for model performance comparison is sourced from \cite{Benato_2022,steinheimer_2021_5710737}. It is a simulated dataset designed to study relativistic nuclear collisions, particularly focusing on the effects of non-equilibrium deconfinement phase transitions. 
\begin{figure}[h]
  \centering
  \includegraphics[scale=0.35]{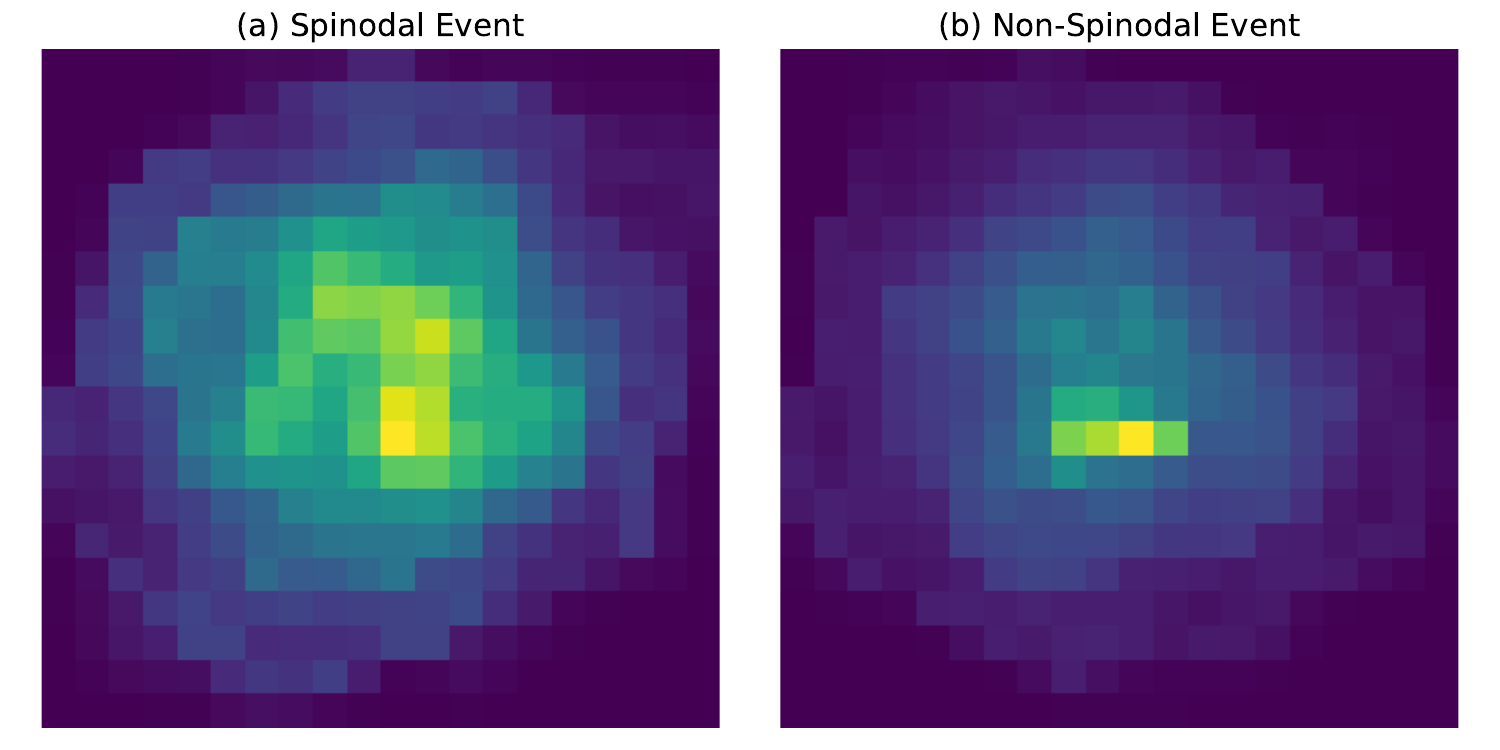}  
  \captionsetup{justification=raggedright, singlelinecheck=false}
 \caption{Spinodal and Non-Spinodal events in the dataset.}
  \label{example}
\end{figure}
The dataset is generated using fluid dynamical simulations that explore the presence of first-order deconfinement phase transitions, considering two scenarios: one with spinodal decomposition and one without. The deep learning models aim to classify these two types of events. The dataset consists of 29,000 central lead-on-lead collision events, each represented as a 20x20 pixel 
(Fig \ref{example})  histogram, which serves as the input for the deep learning models.

\section{Previous Work}
Machine learning models have proven effective in identifying domains and classifying spinodal and non-spinodal events. In \cite{Steinheimer_2019}, machine learning models were employed alongside traditional observables to classify spinodal decomposition in simulated heavy-ion collisions. The study demonstrated that momentum-space features are less informative compared to coordinate-space data when classifying individual events. This finding was further corroborated by \cite{STEINHEIMER2021121867}, which also highlighted the limitations of using momentum space for individual event features. The authors focused on event-averaged observables as a more reliable approach for detecting spinodal events in heavy-ion collisions.

In \cite{Benato_2022}, significant contributions were made to the development of machine learning methods for fundamental physics. The work provides a unified dataset collection and a common interface, facilitating efficient collaborative research in the field. We reference the Convolutional Neural Network (CNN) model presented in \cite{Benato_2022} for the spinodal dataset as a baseline for comparison with the models utilized in this study.

\begin{figure}[h]
  \centering
  \includegraphics[scale=0.4]{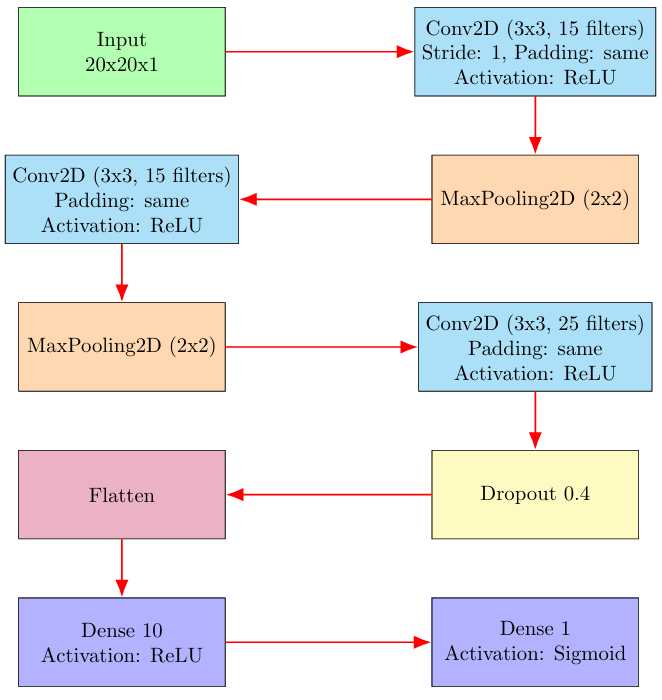}   
  \captionsetup{justification=raggedright, singlelinecheck=false}
  \caption{Architecture diagram of the CNN network.}

  \label{fig1}
\end{figure}

The architecture of the CNN from \cite{Benato_2022} is shown in Fig. \ref{fig1}. It begins with an input layer that performs a 2D convolution using 15 filters, each of size 3x3. These filters scan small patches of the input image with a stride of 1. The activation function used is ReLU, which introduces non-linearity by zeroing out negative values. Following the first convolution layer, there is a max pooling layer with a 2x2 window, which reduces the spatial dimensions, helping the network to focus on important features while reducing computational complexity. 
Next, a second convolution layer, similar to the first, is added to learn more detailed features. After the second pooling layer, a third convolution layer with 25 filters is applied to extract even more complex patterns. Following the convolution layers, a dropout layer with a rate of 0.4 is introduced to prevent overfitting. After the dropout, the model includes a flattening step, followed by a dense layer with 10 neurons and ReLU activation. This dense layer consolidates the features extracted by the convolution layers into a more interpretable form for classification.

The final output is produced by a single neuron with a sigmoid activation function, providing the model's prediction as a probability for the two classes.

\section{Models Description}
In this section, we describe the models used in our work for the classification of the dataset.
\subsection{AlexNet}
We also included AlexNet \cite{alom2018historybeganalexnetcomprehensive} for comparison. AlexNet consists of 5 convolutional layers and 3 fully connected layers, and it played a major role in revolutionizing CNN models. Since AlexNet requires a fixed input size of 224x224 pixels, image preprocessing is necessary when handling images of different sizes.

\subsection{ResNet18}
ResNet, or Residual Network, was introduced to address the problem of vanishing gradients in deep networks. ResNet \cite{he2015deepresiduallearningimage} uses residual connections, also known as skip connections, to bypass layers and mitigate this issue. There are several versions of ResNet, including ResNet18, ResNet34, and ResNet50, where the numerical value indicates the number of layers. In this work, we deployed ResNet18, which includes 8 residual blocks and two 3\( \times 3 \) convolutional layers, followed by residual connections.
\subsection{MobileNet}
MobileNet \cite{howard2017mobilenetsefficientconvolutionalneural} was developed by Google to balance accuracy and efficiency in computer vision tasks on mobile and edge devices. The key feature of MobileNet is its use of depth-wise separable convolutions, which consist of a two-step convolution: depth-wise and point-wise. This design improves computational efficiency.
\subsection{EfficientNet}
EfficientNet \cite{tan2020efficientnetrethinkingmodelscaling}, also introduced by Google, uses a compound scaling method to optimize both accuracy and efficiency. In EfficientNet, the width, depth, and input resolution are scaled uniformly. It has been widely applied in image classification, object detection, and segmentation tasks.
\subsection{MobileViT}
To leverage the strengths of both CNN and Transformer architectures for the classification of the spinodal dataset, we used MobileViT \cite{mehta2022mobilevitlightweightgeneralpurposemobilefriendly}. MobileViT combines convolution layers to extract low-level features with Transformer blocks for capturing global dependencies, followed by additional convolution layers to produce the final output.
\subsection{Neighborhood Attention Transformer (NAT)}
The Neighborhood Attention Transformer (NAT) \cite{hassani2023neighborhoodattentiontransformer} is a neural network architecture designed to improve the efficiency of attention mechanisms in computer vision tasks. Unlike global self-attention, which requires every patch to attend to every other, NAT focuses on local region attention, significantly reducing computational costs.

\subsection{AdaBoost}
AdaBoost \cite{FREUND1997119} is an ensemble technique that combines weak classifiers to form a strong classifier. The contribution of 
each weak classifier to the final output is adjusted based on its performance. Originally designed for binary classification, AdaBoost is well-suited for our analysis.
\begin{figure}[h]
  \centering
  \includegraphics[scale=0.28]{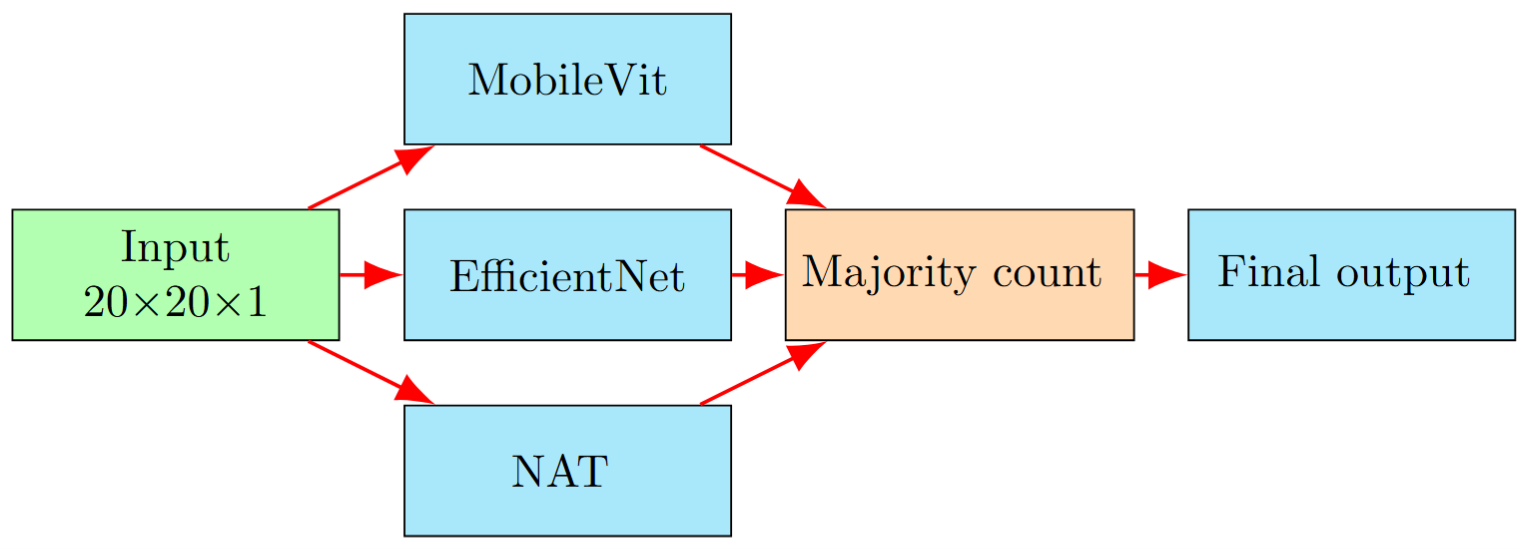}   
  \captionsetup{justification=raggedright, singlelinecheck=false}
  \caption{Majority voting model}
  \label{fig2}
\end{figure}
\subsection{Majority Voting}
In the context of ensemble learning, we developed a model combining NAT, MobileViT, and EfficientNet. The input is fed into all three models as shown in Fig. \ref{fig2}, and each model assigns a class to the event. The final class is then determined through majority voting. The rationale behind this approach is that no single model may capture all patterns or information, so combining models allows us to exploit their collective strengths.

\section{Results and Analysis}
In this section, we discuss the results and analysis of our experiments. The dataset was split into a 70\% training set and a 30\% test set. Additionally, 20\% of the training data was used for validation. 

\begin{figure}[h]
  \centering
  \includegraphics[scale=0.35]{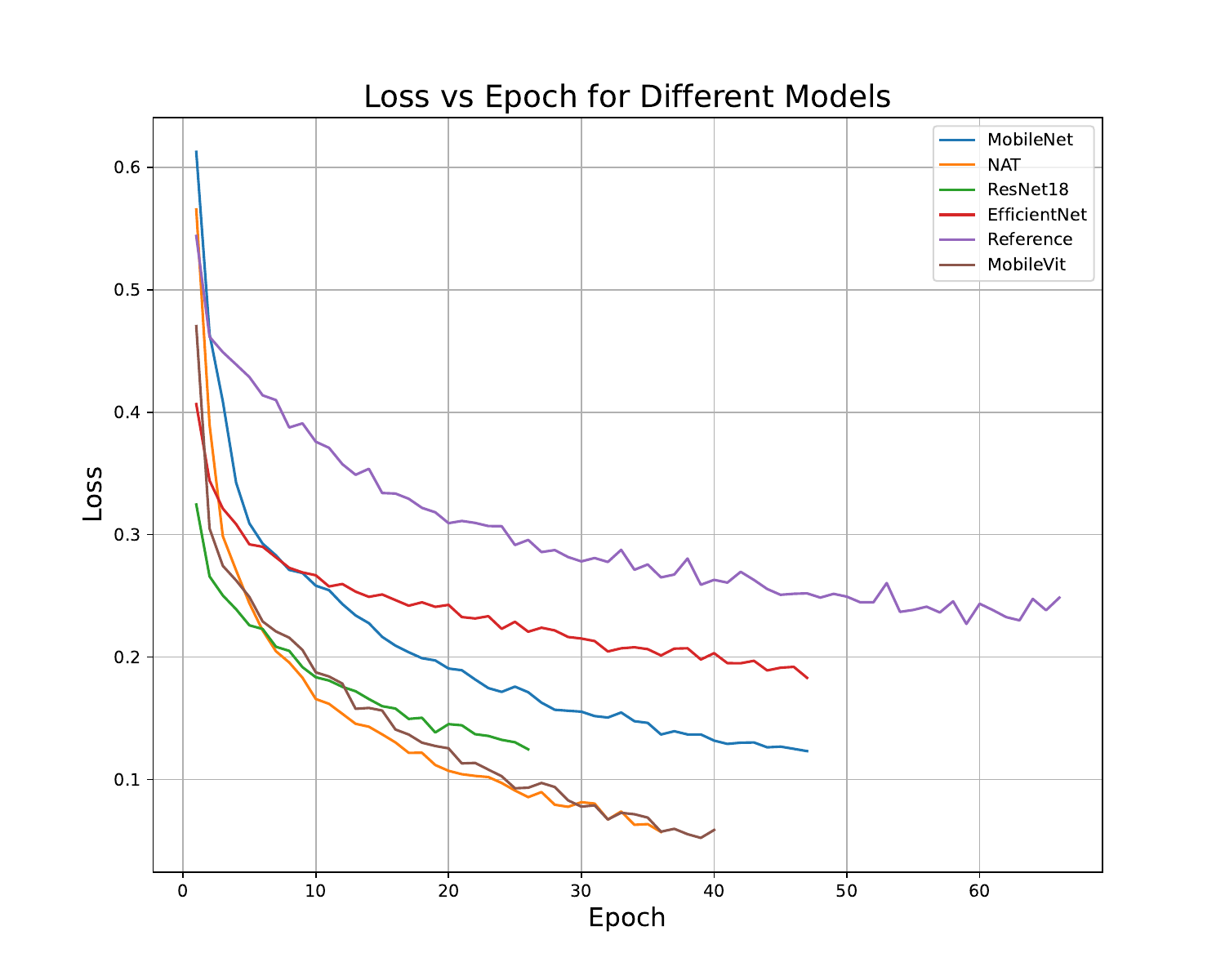}   
  \captionsetup{justification=raggedright, singlelinecheck=false}
  \caption{Loss as a function of number of epochs on the training dataset.}

  \label{fig3}
\end{figure}
We also transformed the images into the CIELAB\cite{ref170} color space to create a transformed dataset, with the aim of exploring whether patterns or features in this transformed space could improve model performance.

Certain models, such as AlexNet, have specific requirements for input image size. In such cases, we modified the network architecture to make it compatible with the standard image size of our dataset.

\begin{figure}[h]
  \centering
  \includegraphics[scale=0.35]{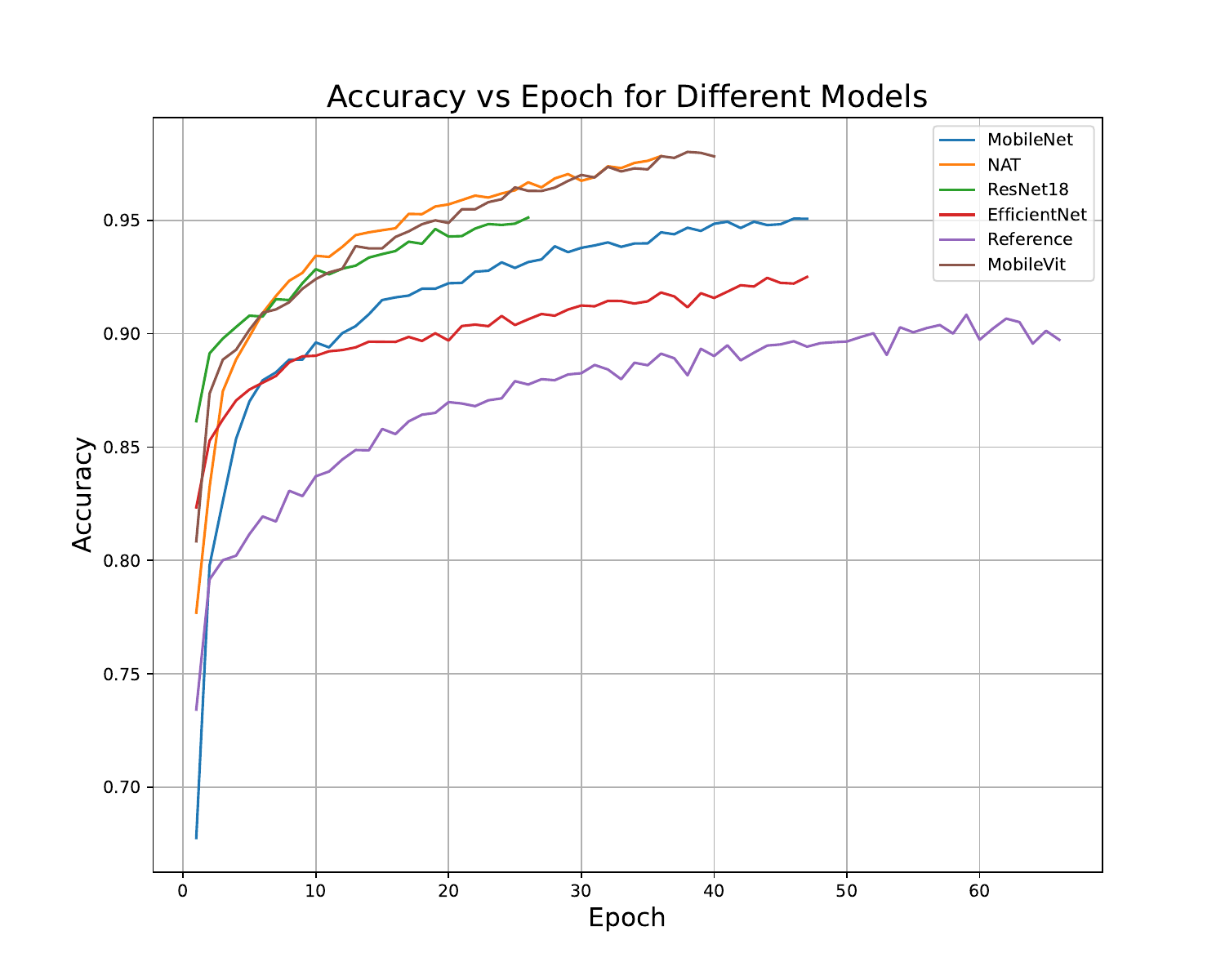}   
  \captionsetup{justification=raggedright, singlelinecheck=false}
  \caption{Accuracy of models as a function of number of epochs on the training dataset.}

  \label{fig4}
\end{figure}

The loss and accuracy for the selected models (those that performed better than the reference model(CNN model) are shown in Fig. \ref{fig3} and Fig. \ref{fig4}. From Fig. \ref{fig3}, we observe that both NAT and MobileViT exhibit similar loss convergence, and they achieve the highest accuracy on the training set, followed by ResNet18 and MobileNet. EfficientNet also performed better than the CNN, though it did not surpass the performance of the other models. 
\begin{figure}[h]
  \centering
  \includegraphics[scale=0.35]{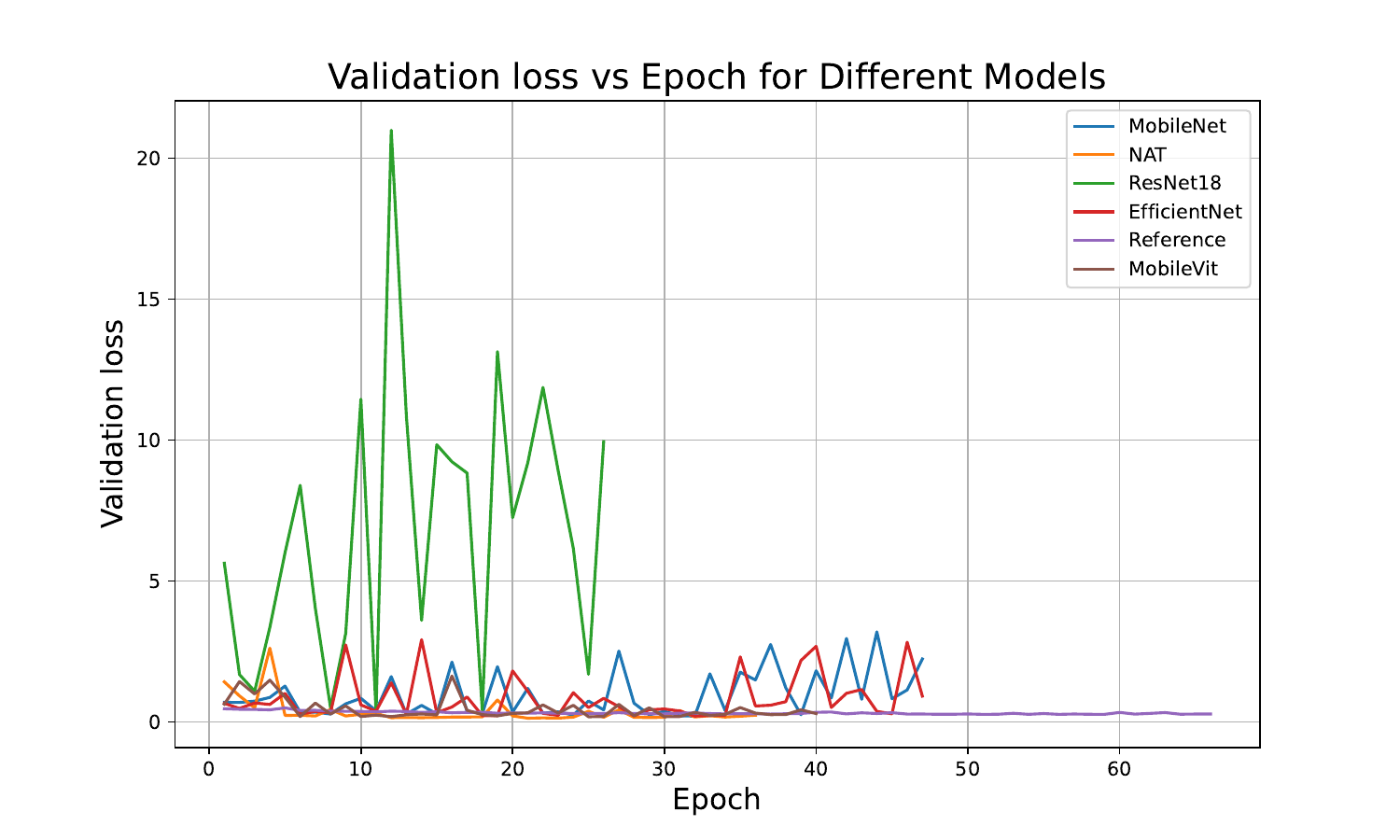}   
  \captionsetup{justification=raggedright, singlelinecheck=false}
  \caption{Validation loss during training.}

  \label{fig5}
  \end{figure}
  The models shown in Fig. \ref{fig3} and Fig. \ref{fig4} have been modified as discussed earlier to accommodate the dataset's input size requirements. The validation loss and validation accuracy are shown in Fig. \ref{fig5} and Fig. \ref{fig6}, respectively. The most fluctuation in validation loss was observed in ResNet18, indicating that it may not generalize well to test data, a prediction confirmed by its performance on the test set. Both NAT and MobileViT exhibit stable trends similar to the CNN model.
  \begin{figure}[h]
  \centering
  \includegraphics[scale=0.35]{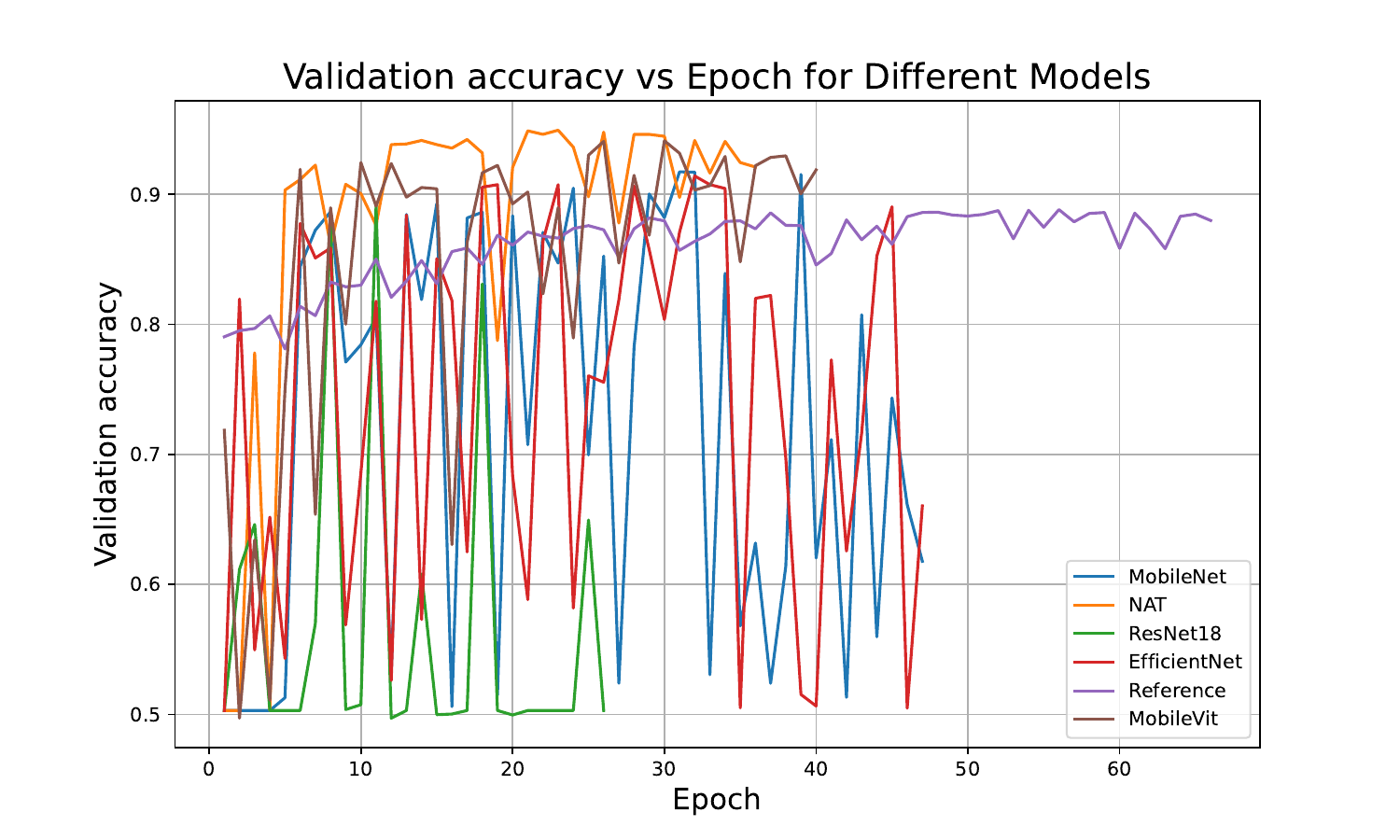}   
  \captionsetup{justification=raggedright, singlelinecheck=false}
 \caption{Validation accuracy during training.}
 \label{fig6}
\end{figure}

In terms of validation accuracy, NAT and MobileViT outperform the other models. While most models (excluding the CNN) show significant fluctuations in validation accuracy, NAT and MobileViT stabilize after a certain number of epochs, reflecting their superior generalization capabilities.

\begin{figure}[h]
  \centering
  \includegraphics[scale=0.35]{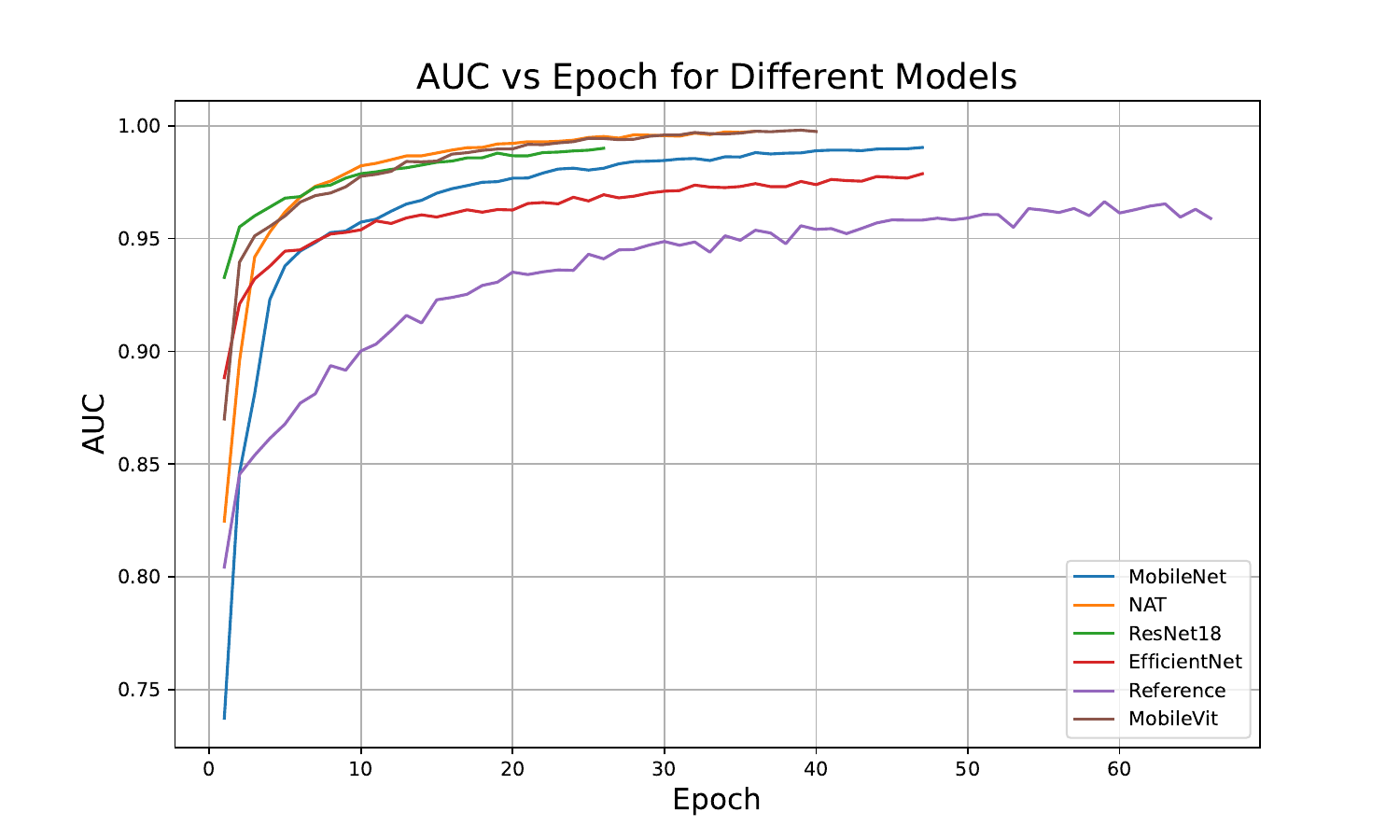}   
  \captionsetup{justification=raggedright, singlelinecheck=false}
  \caption{Area under the curve (AUC) on the training dataset.}
 \label{fig7}
 \end{figure}
Fig. \ref{fig7} provides an AUC (Area Under the Curve) comparison for various models during training. AUC measures the true positive rate relative to the false positive rate at various classification thresholds. An AUC close to 1 indicates that a model is highly effective at distinguishing between the two classes. Once again, NAT achieves the highest AUC, followed by MobileViT and ResNet18, while the reference model shows the lowest AUC on the training dataset.

Considering the overall picture, our findings suggest that NAT and MobileViT should deliver the best performance on test data among all the models considered. They demonstrate the highest AUC, better validation and training accuracy, as well as the lowest validation and training loss. This is indeed the case, as summarized in Table \ref{Table1}, where NAT and MobileViT achieve the best overall performance.

On the other hand, AdaBoost did not prove useful for our dataset, performing worse than the CNN model. However, our ensemble model using majority voting outperforms the CNN and even surpasses some individual networks. AlexNet also performed slightly better than the CNN. 
\begin{table}[h!]
    \centering
      \caption{Results on the test dataset for various models.}
    \begin{tabular}{c c c c}
        \hline
        Model & Accuracy(\%) & AUC & F1 Score \\ \hline \hline 
        Reference Model & 0.884 & 0.95 & 0.88 \\ \hline 
        Adaboost & 0.801 & 0.86 & 0.78 \\  \hline
        Majority Voting & 0.943 & 0.94 & \textbf{0.94} \\  \hline
        MobileVit & 0.942 & \textbf{0.98} & \textbf{0.94} \\  \hline
        NAT & \textbf{0.946} & \textbf{0.98} & \textbf{0.94}\\ \hline
        EfficientNet & 0.910 & 0.97 & 0.91\\ \hline
       
        AlexNet & 0.910 & 0.97 & 0.90\\ \hline
      
        ResNet18 & 0.890 & 0.96 & 0.88\\ \hline
      
        MobileNet & 0.930 & 0.98 & 0.93 \\ \\ \hline
        CIELAB dataset \\ \hline
        \hline 
        MobileVit &  0.941 & 0.98 & 0.94 \\ \hline
        NAT &  0.940 & 0.98 & 0.94 \\ 
        \hline 
        Efficient & 0.910 & 0.97 & 0.91\\ \hline
        MobileNet & 0.916 & 0.97 & 0.91\\ \hline

    \end{tabular}
        \label{Table1}
\end{table}
When we provided the transformed dataset (CIELAB) as input, the performance was close to that of the original dataset, except in the case of MobileNet, where it decreased. Therefore, transforming to CIELAB did not help us in this case.
Next, we examine failure cases. Table \ref{Table2} lists the misclassified instances for the top three models: NAT, MobileViT, and EfficientNet. Interestingly, NAT and the reference CNN share 4.6\% of common misclassified cases. 
\begin{table}[h!]
    \centering
     \caption{Table showing common  misclassified cases.}
    \begin{tabular}{c c c c}
    \hline 
        Model & Misclassified (\%) \\ \hline \hline
       CNN and NAT & 4.6   \\ \hline 
       CNN and MobileVit &  3.5 \\ \hline 
       CNN and EfficientNet & 5.2  \\ \hline 
       NAT and MobileVit  &  3.4  \\ \hline 
    \end{tabular}
  \label{Table2}
\end{table}
NAT and MobileViT, on the other hand, share 3.5\% of misclassified cases, which constitutes 66\% of NAT’s misclassifications and 60\% of MobileViT’s. Thus, more than half of the misclassified cases are common to both NAT and MobileViT. Note that the fewer the uncommon cases we have, the better our majority voting model will perform.

\section{Conclusion}
Our findings conclude that deep learning models such as NAT and MobileViT offer strong accuracy for the classification of the spinodal dataset. Future work could focus on employing larger datasets and exploring more advanced architectures to further improve performance. Additionally, preprocessing techniques applied to the images before feeding them into deep networks may enhance classification accuracy, ultimately contributing to a deeper understanding of nuclear matter dynamics in heavy-ion collisions.

\bibliographystyle{IEEEtran} 
\bibliography{Mybib.bib}

\end{document}